# Off-line Constraint Propagation for Efficient HPSG Processing

## Walt Detmar Meurers and Guido Minnen*

SFB 340, Kleine Wilhelmstr. 113, 72074 Tübingen, Germany
E-mail: {dm,minnen}@sfs.nphil.uni-tuebingen.de

## Abstract

We investigate the use of a technique developed in the constraint programming community called *constraint propagation* to automatically make a HPSG theory more specific at those places where linguistically motivated underspecification would lead to inefficient processing. We discuss two concrete HPSG examples showing how off-line constraint propagation helps improve processing efficiency.

## 1 Introduction

A major goal of a linguist writing HPSG theories is to express very general constraints to capture linguistic phenomena, leaving as much as possible underspecified. When such a HPSG theory is implemented faithfully, either processing is inefficient because only little information is available to guide the constraint resolution process, or the linguistic theory is annotated with information to guide processing. Usually such annotations are provided manually – a very time consuming and error-prone process which can change the original linguistic theory. In this paper we show that it is possible to automatically make a theory more specific at those places where linguistically motivated underspecification would lead to inefficient processing.

An off-line compilation technique called *constraint propagation* is used to improve processing efficiency by means of propagating constraints already expressed in the theory. Programs do not necessarily profit from constraint propagation. For processing grammars, constraint propagation can be very useful, since it makes it possible to process the general constraints expressing linguistic generalizations specified by the linguist, without falling prey to massive nondeterminism. The relevant observation here is that even though certain places in a grammar are underspecified, the grammar does contain enough constraining information – it just needs to be moved to guide processing. Constraint propagation also makes it possible to advance automatically generated encodings, such as, for example, the definite clause encoding of HPSG grammars introduced in Götz and Meurers (1995).

Constraint propagation can be performed on-line as in Le Provost and Wallace (1993) or it can be used to make programs more specific through off-line compilation as in Marriott, Naish, and Lassez (1988). In this paper we will focus on the off-line application of constraint propagation. While on-line constraint propagation is more space efficient since information in the code does not need to be duplicated, the off-line process can relieve the run-time from significant overhead.[1] We conjecture that the time-space tradeoff can be exploited by doing off-line constraint propagation only selectively. This presupposes that the places in a program which will profit from constraint propagation can be automatically located by abstract interpretation. An investigation of such an abstract interpretation method, however, is beyond the scope of this paper.

---

The research reported here was sponsored by Teilprojekt B4 'From Constraints to Rules: Efficient Compilation of HPSG Grammars' of SFB 340 'Sprachtheoretische Grundlagen für die Computerlinguistik' of the Deutsche Forschungsgemeinschaft. The authors wish to thank Thilo Götz, Dale Gerdemann and the anonymous reviewers for comments and discussion. Of course, the authors are responsible for all remaining errors.

*The authors are listed alphabetically. URL: http://www.sfs.nphil.uni-tuebingen.de/sfb4

[1]In case an operation for "subtraction" is available for the data structure used, it may be possible to reduce the space cost of the off-line process by eliminating the propagated constraints from their original specification site.

Other techniques to prune the search space that are used in practical natural language processing are *dynamic coroutining*, also referred to as (goal) freezing or delaying, and *static coroutining* by means of Unfold/Fold transformation (Tamaki and Sato, 1984). It is important to differentiate between coroutining and constraint propagation: Coroutining changes the way in which the search space is investigated by moving goals through a grammar either on- or off-line. Constraint propagation as conceived in this paper reduces the search space by making the arguments of calls to goals more specific. As we will discuss in section 3, a combination of both techniques can be very useful as constraint propagation can be used to extract restricting information from the definition of goals also in cases where freezing of the call to these goals would hide this information.

This paper is organized as follows: We start with a discussion of two concrete HPSG examples showing how constraint propagation helps improve processing efficiency (sections 2 and 3). In section 4 several implementations of constraint propagation algorithms are discussed. Finally, in section 5 we provide some implementation results.

## 2 Efficient processing of ID Schemata

In lexically oriented grammar formalisms like HPSG, the ID schemata specified by the linguist are very schematic since much syntactic information is specified in the lexicon. In faithful implementations this leads to inefficiency in top-down processing because it often is no longer possible to detect locally whether an ID schema applies or not. Consider, for example, the head-adjunct schema and the head-specifier schema of HPSG in the figures 1 and 2.[2] Due to underspecification, it cannot be determined locally whether the head-adjunct schema can expand specifiers or not. Only upon lexical lookup is it revealed that the head-adjunct schema does not have to be considered for specifiers: The lexicon contains only lexical entries like the one sketched in figure 3, which specify the category they modify to have a *substantive* head, in this case a *noun*. This specification will therefore always clash with the specification in the head-specifier schema which demands a *functional* head value for the specifier daughter.

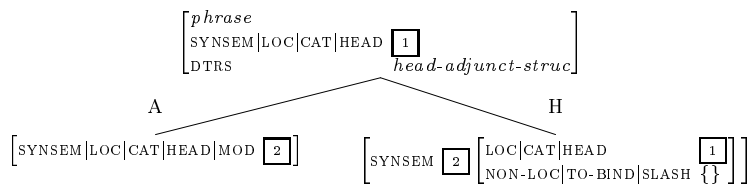

Figure 1: *The Head-Adjunct ID Schema from Pollard and Sag (1994).*

The sketched efficiency problem seems to suggest that top-down processing is not the right processing strategy to adopt for processing of lexically oriented grammar formalisms. This, however, is not necessarily the case. Strict bottom-up processing means that no filtering information resulting from the start category is made available. To have some guiding information in the case of parsing an extra-logical treatment of the input string can be used. However, it is unclear what such a treatment should look like for theories using more elaborate linearization operations. Furthermore, refraining from taking into account information provided by the start category is virtually impossible in the case of generation and it is unclear what an extra-logical treatment of the logical form in a similar fashion as in parsing could look like. There exists an off-line compilation technique called *magic* that allows for filtering given a strict bottom-up processing strategy.[3] However, processing of magic compiled grammars suffers from linguistically motivated underspecification as discussed above just the same.

---

[2]The figures show the Head-Adjunct Schema as expressed in the appendix of Pollard and Sag (1994) and the Head-Specifier Schema from chapter 9 of the same book - both including the effect of the Head Feature Principle.

[3]See among others, Ramakrishnan (1991). In Minnen (1996) applications of these techniques to natural language processing are discussed.

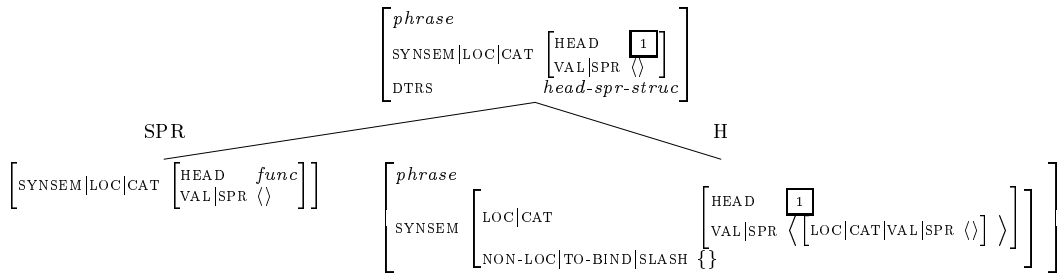

Figure 2: *The Head-Specifier ID Schema from Pollard and Sag (1994).*

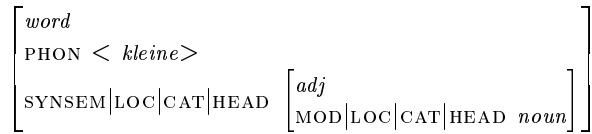

Figure 3: *The lexical entry for the adjective 'kleine'.*

Returning to the above example, the insight behind constraint propagation is that lifting the common restricting information contained in the lexical entries up into the head-adjunct schema makes it possible to determine locally that there are no modified specifiers in the grammar. In other words, applying constraint propagation to the head-adjunct schema of figure 1 in a grammar with a lexicon in which the only modifying entries select *substantive* heads, propagates the constraint [SYNSEM|LOC|CAT|HEAD *subst*] into the mother of the head-adjunct schema. The resulting head-adjunct schema shown in figure 4 is now specific enough to convey immediately that it cannot be used when specifiers need to be licensed.

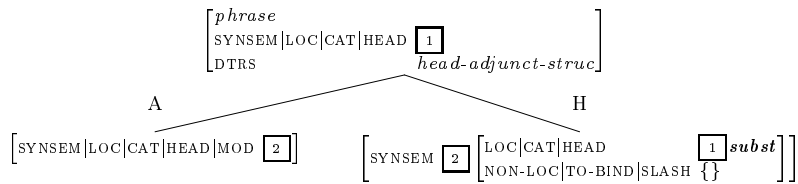

Figure 4: *The Head-Adjunct ID Schema after constraint propagation.*

Note that this way of making grammars more specific is an off-line process performed completely automatically. It allows the grammar writer to specify theories in a lexically oriented fashion without any additional procedural specifications.

# 3 Efficient processing of the lexicon

Constraint propagation can also be applied to optimize automatically generated lexica. In Meurers and Minnen (1995) a compiler is described which translates a set of HPSG lexical rules and their interaction into definite relations used to constrain lexical entries. This, so-called, *covariation approach* uses the generalizations captured by lexical rules for processing and makes it possible to deal with the infinite lexica proposed in many recent HPSG theories.

The linguist inputs the lexical rules used in his/her theory. On the basis of this specification and the signature of the proposed grammar, the covariation compiler automatically deduces the transfer of properties which were left unspecified in the lexical rule provided by the linguist. The compiler then uses the lexical rules and lexical entries to produce a definite clause encoding of lexical rules and their possible interaction. The resulting lexicon consists of extended lexical entries calling an interaction predicate encoding the entries which can be derived by lexical rule applications. Figure 5 shows an example for an extended lexical entry: a simplified entry for a German auxiliary using argument raising in the style of Hinrichs and Nakazawa (1989). The call to the interaction predicate encodes the possible

$$
\text{extended\_lex\_entry(}\boxed{\text{OUT}}\text{):- interaction\_0(}
\begin{bmatrix}
\text{PHON} & <k\ddot{o}nnen> \\
\text{VFORM} & bse \\
\text{SUBCAT} & < \begin{bmatrix} \text{VFORM} & bse \\ \text{SUBCAT} & \boxed{1} \\ \text{CONT} & \boxed{2} \end{bmatrix} \mid \boxed{1} > \\
\text{CONT} & \begin{bmatrix} k\ddot{o}nnen' \\ \text{ARG} & \boxed{2} \end{bmatrix}
\end{bmatrix}
\text{, }\boxed{\text{OUT}}\text{).}
$$

Figure 5: *The extended lexical entry for the modal auxiliary 'können' (can).*

sequences of lexical rule applications. For a simple theory with a Complement Extraction Lexical Rule (CELR) and a Finitivization Lexical Rule (FINLR) the slightly simplified interaction predicate looks as shown in figure 6.[4] The encoding in figure 6 already contains the deduced transfer information in the

$$
\text{interaction\_0(}\boxed{\text{IN}}
\begin{bmatrix}
\text{PHON} & \boxed{1} \\
\text{VFORM} & \boxed{2} \\
\text{CONT} & \boxed{3}
\end{bmatrix}
\text{, }\boxed{\text{OUT}}\text{):- celr(}\boxed{\text{IN}}\text{, }\boxed{\text{AUX}}\text{), interaction\_0(}\boxed{\text{AUX}}
\begin{bmatrix}
\text{PHON} & \boxed{1} \\
\text{VFORM} & \boxed{2} \\
\text{CONT} & \boxed{3}
\end{bmatrix}
\text{, }\boxed{\text{OUT}}\text{).}
$$

$$
\text{interaction\_0(}\boxed{\text{IN}}
\begin{bmatrix}
\text{PHON} & \boxed{1} \\
\text{SUBCAT} & \boxed{2} \\
\text{SLASH} & \boxed{3} \\
\text{CONT} & \boxed{4}
\end{bmatrix}
\text{, }\boxed{\text{OUT}}\text{):- finlr(}\boxed{\text{IN}}\text{, }\boxed{\text{AUX}}\text{), interaction\_1(}\boxed{\text{AUX}}
\begin{bmatrix}
\text{PHON} & \boxed{1} \\
\text{SUBCAT} & \boxed{2} \\
\text{SLASH} & \boxed{3} \\
\text{CONT} & \boxed{4}
\end{bmatrix}
\text{, }\boxed{\text{OUT}}\text{).}
$$

$$
\text{interaction\_0(}\boxed{\text{OUT}}\text{, }\boxed{\text{OUT}}\text{).}
$$
$$
\text{interaction\_1(}\boxed{\text{OUT}}\text{, }\boxed{\text{OUT}}\text{).}
$$

$$
\text{celr(}
\begin{bmatrix}
\text{VFORM} & bse \\
\text{SUBCAT} & < \boxed{1} \mid \boxed{2} > \\
\text{SLASH} & \boxed{3}
\end{bmatrix}
\text{, }
\begin{bmatrix}
\text{SUBCAT} & \boxed{2} \\
\text{SLASH} & < \boxed{1} \mid \boxed{3} >
\end{bmatrix}
\text{).}
$$

$$
\text{finlr(}
\begin{bmatrix}
\text{PHON} & \boxed{1} \\
\text{VFORM} & bse
\end{bmatrix}
\text{, }
\begin{bmatrix}
\text{PHON} & \boxed{2} \\
\text{VFORM} & fin
\end{bmatrix}
\text{) :- third\_fin(}\boxed{1}\text{, }\boxed{2}\text{).}
$$

third_fin(können,kann).
...

Figure 6: *Encoding sequences of lexical rule application by means of definite relations.*

call to the lexical rule predicates; for example, the PHON, VFORM, and CONT value is transferred for the CELR by adding the corresponding structure sharings to the $\boxed{\text{IN}}$ and $\boxed{\text{AUX}}$ tags which also appear in the call to the celr/2 predicate.

The automatically obtained encoding of lexical rule application in lexical entries shown in the above figures is not very efficient since before execution of the call to the interaction predicate it is unknown which information of the base lexical entry ends up in a derived lexical entry. One is therefore forced to execute the call to the interaction predicate directly when the lexical entry is used during processing, independent of the processing strategy used. Otherwise there is no information available to restrict the search space of a generation or parsing process.

---

[4] The lexical rules in figure 6 are simplified versions of the CELR (Pollard and Sag, 1994, p. 378) and the Third-Singular Inflectional Rule (Pollard and Sag, 1987, p. 210).

Off-line constraint propagation can be used to avoid this by factoring out the information which is common to all solutions for the called interaction predicate. This is accomplished by computing the most specific generalization of these solutions and lifting this common information into the extended lexical entries. Let C be the common information, and $D_1, \ldots, D_k$ the solutions for the interaction predicate called. Then by distributivity we factor out C in $(C \wedge D_1) \vee \ldots \vee (C \wedge D_k)$ to obtain $C \wedge (D_1 \vee \ldots \vee D_k)$, where the D are assumed to contain no further common factors. The result of performing constraint propagation on the extended lexical entry for 'können' is given in figure 7. In the next section we investigate in more detail how this result is achieved.

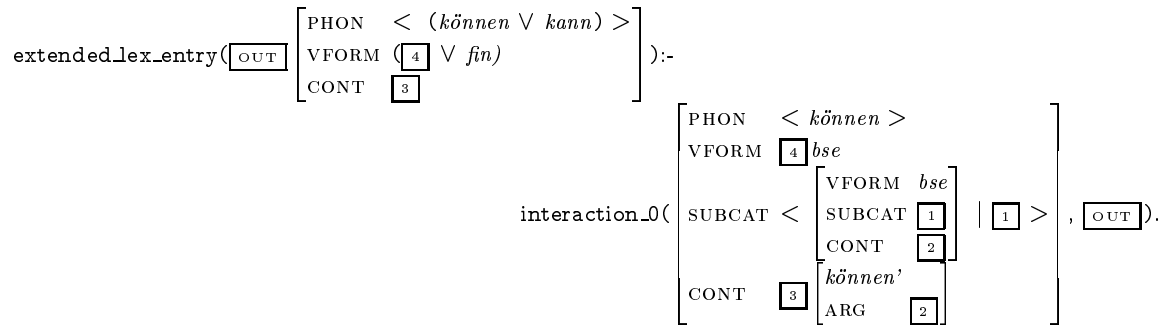

Figure 7: *The extended lexical entry for 'können' after specialized constraint propagation.*

Delaying the call to an interaction predicate as in Van Noord and Bouma (1994) by freezing the recursive application of a lexical rule on the basis of user-specified delay information, can hide important restricting information because it is specified in the definition of the frozen goal. Therefore constraint propagation can be useful, also when coroutining techniques are used.

As discussed in Griffith (1996) an extension of the constraint language with *contexted constraints*, also referred to as dependent or named disjunctions, in certain cases makes it possible to circumvent constraint propagation. Encoding the disjunctive possibilities for lexical rule application using contexted constraints instead of definite clause attachments makes all relevant linguistic information available at lexical look-up. In case of infinite lexica, though, a definite clause encoding of disjunctive possibilities is still necessary and constraint propagation is indispensable for efficient processing (see section 5).

# 4   Implementing Constraint Propagation

In this section we discuss implementations of some constraint propagation algorithms (in Prolog). We first present constraint propagation using a simple top-down interpreter and point out the problems of this basic algorithm. Subsequently, possible extensions of this interpreter with a, so-called, *branch-and-bound* optimization (Le Provost and Wallace, 1993) and a depth-bound are discussed. Finally, we show that it is possible to use knowledge about the specific structure of certain encodings to obtain specialized constraint propagation algorithms. In our case, we can exploit our knowledge of the encoding of the lexicon produced by the lexical rule compiler to define a specialized top-down interpreter that relieves us from termination problems related to the covariation encoding of infinite lexica.

## 4.1   Top-down constraint propagation

Consider the predicate `constraint_propagation_on_goal/0` in figure 8. The predicate `get_goal/0` gets a particular goal on which we want to perform constraint propagation.[5] Subsequently, `generalized_solutions_for_goal/2` is called to produce a possibly more specific instance of this goal. The call to `write_goal/1` replaces the original goal with the possibly more specific goal obtained. As shown in figure 9 `generalized_solutions_for_goal/2` computes an instance `GeneralizedSolutionsForGoal` of `Goal` by finding all its solu-

---

```
constraint_propagation_on_goal:-
    get_goal(Goal),
    generalized_solutions_for_goal(Goal,MoreSpecificGoal),
    write_goal(MoreSpecificGoal).
```

Figure 8: *A predicate defining simple off-line constraint propagation on a goal.*

tions with a call to `top_down_interpret/1` and subsequently generalizing over all the solutions. Figure 10

```
generalized_solutions_for_goal(Goal,GeneralizedSolutionsForGoal):-
    findall( Goal,
             top_down_interpret(Goal),
             SolutionList
           ),
    generalize_all_solutions(SolutionList,GeneralizedSolutionsForGoal).
```

Figure 9: *Generalizing all solutions for goal.*

provides the definition of `top_down_interpret/1`, a top-down interpreter taken from Pereira and Shieber (1987, pp. 160f.).[6] This interpreter falls prey to nontermination. For example, in the case of the recursive

```
top_down_interpret(true).
top_down_interpret(Goal):-
    clause((Goal :- Body)),
    top_down_interpret(Body).
top_down_interpret((Body1, Body2)):-
    top_down_interpret(Body1),
    top_down_interpret(Body2).
```

Figure 10: *A simple top-down interpreter.*

CELR of figure 6 it is possible to remove elements from a (subcategorization) list that is underspecified as in the extended lexical entry of figure 5 over and over again.

Motivated by efficiency considerations, Le Provost and Wallace (1993) propose the *branch-and-bound* optimization. This optimization also improves termination behaviour. However, there exist linguistically motivated types of recursion for which branch-and-bound does not terminate either. Minnen et al. (1996) introduce the notion of a *building series*. Intuitively understood, a building series "builds up" a structure recursively until it matches a "base" case.[7] This type of recursion is problematic for top-down processing as this building can go on forever. Branch-and-bound does not ensure termination in the light of this type of recursion.

These termination problems necessitate an alternative implementation that avoids infinite loops. One possibility is to extend the interpreter in figure 10 with a depth-bound as shown in figure 11.[8] Notice that the use of this highly incomplete interpreter for constraint propagation can only lead to a common factor that is to general. Intuitively understood, the depth-bound can only cut off branches of the search

---

[6]The predicate is renamed here for expository reasons. Nonunit and unit clauses serving as data for the interpreter are represented as `clause(( Head :- Body ))`. and `clause(( Head :- true ))`., respectively.

[7]An example of a lexical rule that exhibits this type of recursion on structural information is the Add Adjuncts Lexical Rule proposed in Van Noord and Bouma (1994).

[8]The call to `top_down_interpret/3` in `generalize_solutions_for_goal/2` shown in figure 9 has to be changed accordingly.

```
db_top_down_interpret(true, Depth, Max):-
    Depth < Max.
db_top_down_interpret(Goal, Depth, Max):-
    Depth < Max,
    clause((Goal :- Body)),
    NewDepth is Depth + 1,
    db_top_down_interpret(Body, NewDepth, Max).
db_top_down_interpret((Body1, Body2), Depth, Max):-
    Depth < Max,
    db_top_down_interpret(Body1, Depth, Max),
    db_top_down_interpret(Body2, Depth, Max).
db_top_down_interpret(_Goal, _Depth, _Max):-
    Depth > Max.
```

Figure 11: *A depth-bounded top-down interpreter.*

space which will eventually fail or lead to a solution more specific than the partial solution that has been computed. When the depth-bound hits clause 4 of db_top_down_interpret/3 in figure 11, the result returned in the first argument does not become further instantiated. As a result the MoreSpecificGoal computed can never become too specific and correctness is guaranteed.

While the depth-bounded interpreter can be employed in general, it is far from optimal to use it for constraint propagation of the covariation encoding of the lexicon. This is due to the fact that lexical rule application is encoded as forward chaining using *accumulator passing* (O'Keefe, 1990): The `out` argument of an interaction predicate gets instantiated upon hitting a base case, i.e., a unit interaction clause. It serves only to "return" the lexical entry eventually derived. When the depth-bound cuts off a particular branch of the search space that corresponds to a recursively defined interaction predicate, the `out` argument remains completely uninstantiated. Consequently, generalizing over all possible (partial) solutions does not lead to a common factor that is more specific than the original goal selected by get_goal/1. In the next section, we show that it is possible to overcome this problem with a specialized interpreter.

## 4.2 Specialized Constraint Propagation

We employ a specialized top-down interpreter that allows us to extract an informative common factor using constraint propagation even in cases of a covariation encoding of an infinite lexicon. The specialized interpreter makes the use of a depth-bound to ensure termination of the interpretation of the interaction predicates superfluous.[9] Intuitively understood, the specialized interpreter exploits the fact that automatic property transfer is not influenced by recursion. I.e., the specifications that are left unchanged by a recursive lexical rule are independent of the number of times the rule is applied.

We discuss a possible extension of the simple top-down interpreter given in figure 10. For expository reasons the interpreter given in figure 12 is simplified in the sense that it deals only with directly recursive interaction predicates such as the one given in figure 6. Indirectly recursive interaction predicates necessitate a further extension of the interpreter with a tabelling technique as indirect recursion can not be identified locally, i.e., as a property of the interaction clause under consideration. The original top-down interpreter is extended with an extra clause, i.e., the second clause of spec_top_down_interpret/1, which is specialized to deal with recursive interaction predicates which are identified by means of a call to recursive_interaction_clause/1. By eliminating the call to the lexical rule predicate (corresponding to the application of the recursive lexical rule) the interpreter abstracts over the information that is changed by the recursive lexical rule. As a result, only unchanged information remains. Subsequently, spec_top_down_interpret/2 is called to ensure that the same recursive interaction predicate is not called

---

[9]As nontermination can not only result from recursive interaction predicates, a depth-bound might still be needed for the other predicates. We ignore this complication in the remainder of this section for expository reasons.

(over and over) again.[10]

```
spec_top_down_interpret(true).
spec_top_down_interpret(Goal):-
    clause((Goal :- Body)),
    recursive_interaction_clause((Goal :- Body)),
                        % True if the retrieved clause is a directly recursive
                        % interaction clause.
    make_body_more_general(Body, AdaptedBody),
                        % Removes the call to the recursive lexical rule predicate from
                        % Body in order to abstract over changed information.
    spec_top_down_interpret(AdaptedBody,(Goal :- Body)).
spec_top_down_interpret(Goal):-
    clause((Goal :- Body)),
    \+ recursive_interaction_clause((Goal :- Body)),
    spec_top_down_interpret(Body).
spec_top_down_interpret((Body1, Body2)):-
    spec_top_down_interpret(Body1),
    spec_top_down_interpret(Body2).

spec_top_down_interpret(Goal, RecursiveInteractionCLause):-
    clause((Goal :- Body)),
    \+ (Goal :- Body) = RecursiveInteractionCLause,
                        % Avoid repeated application of RecursiveInteractionCLause.
    spec_top_down_interpret(Body).
```

Figure 12: *A top-down interpreter specialized for constraint propagation on (calls to) interaction predicates in a covariation lexicon.*

Since we abstract over the information that is changed by a recursive lexical rule the common factor that is extracted by means of performing constraint propagation with the specialized top-down interpreter might be too general: In case we are dealing with an infinite lexicon not all (possible infinite) applications of a recursive lexical rules are performed and there might be cases in which the application of a lexical rule after the n-th cycle is impossible even though we are taking it into account during constraint propagation. It is important to note though that such a situation can only lead to a common factor that is too general since generalizing over too large a set of solutions can only lead to a less specific generalization, not a more specific one. Therefore constraint propagation does not influence the soundness and completeness of the encoding. At run-time the additional lexical rule applications not ruled out by constraint propagation will simply fail.

Reconsider the definite clause encoding in figure 6. As a result of the fact that repeated recursive application of `interaction_0/2` is avoided, much relevant information can be lifted into the extended lexical entry. Figure 7 given in the previous section shows the result of performing specialized constraint propagation to the lexical entry for 'können' (figure 5).

## 4.3   Constant time lexical lookup

As figure 7 shows, optimizing the extended lexical entries by means of specialized constraint propagation can also lift up phonological information in case of infinite lexica.[11] This information can be used to index the lexicon so that constant time lexical lookup can be achieved. For this purpose, the extended

---

[10]We exploit the fact that two interaction clauses can never stand in the subsumption relation. Otherwise, a more elaborate equality test is needed in `spec_top_down_interpret/2` to avoid repeated application.

[11]If there are recursive phonology changing rules the phonological information cannot be lifted by the constraint propagation using the specialized interpreter presented.

lexical entry is split up as shown in figure 13. On the basis of the input string it is now possible to

```
indexed_lex_entry(können,[OUT] [PHON < können >] ):- extended_lex_entry([OUT]).
indexed_lex_entry(kann,[OUT] [PHON < kann >] ):- extended_lex_entry([OUT]).
...
```

Figure 13: *The result of splitting up the optimized lexical entry in figure 7.*

access the lexicon in constant time. Without specialized constraint propagation this is impossible as the possible values of the phonology feature are hidden away deep in the covariation encoding of the lexical entries that can be derived from the base lexical entry.

# 5 Implementation Results

The depth-bounded constraint propagation method was implemented for the ConTroll system (Gerdemann and King, 1994; Götz and Meurers, 1995 and 1996) under Prolog. Test results on a complex grammar implementing an analysis of partial VP topicalization in German (Hinrichs, Meurers, and Nakazawa, 1994) show that constraint propagation significantly improves parsing with a covariation encoding of lexical rules. For the lexica produced by the covariation compiler, the implementation revealed that the most specific generalization which is propagated contains much valuable information. This is the case because usually the lexical entries resulting from lexical rule application only differ in few specifications compared to the number of specifications in a base lexical entry. The relation between parsing times with the expanded (EXP), the covariation (COV) and the constraint propagated covariation (OPT) lexicon for the above grammar can be represented as OPT : EXP : COV = 1 : 1.3 : 14.

Many of the referenced papers by the authors and their colleagues are available electronically over the URL given on the first page.